# Cue-word Driven Neural Response Generation with a Shrinking Vocabulary

**Qiansheng Wang[2], Yuxin Liu[2], Chengguo Lv[2], Zhen Wang[2]
and Guohong Fu[1]**
1. Institute of Artificial Intelligence, Suchow University
2. School of Computer Science and Technology, Heilongjiang University
{chncwang,yuhsin.lyx}@gmail.com, 2004085@hlju.edu.cn,
wangzhnlp@163.com, ghfu@suda.edu.cn

## Abstract

Open-domain response generation is the task of generating sensible and informative responses to the source sentence. However, neural models tend to generate safe and meaningless responses. While cue-word introducing approaches encourage responses with concrete semantics and have shown tremendous potential, they still fail to explore diverse responses during decoding. In this paper, we propose a novel but natural approach that can produce multiple cue-words during decoding, and then uses the produced cue-words to drive decoding and shrinks the decoding vocabulary. Thus the neural generation model can explore the full space of responses and discover informative ones with efficiency. Experimental results show that our approach significantly outperforms several strong baseline models with much lower decoding complexity. Especially, our approach can converge to concrete semantics more efficiently during decoding.

## 1 Introduction

In recent years, neural response generation has achieved significant success in both academic (Shang et al., 2015; Zhang et al., 2019; Adiwardana et al.,2020; Roller et al., 2020) and commercial worlds (Li et al., 2017; Zhou et al., 2018). Text generation approaches in these studies are borrowed from neural translation (Bahdanau et al., 2014), which uses the sequence-to-sequence(seq2seq) model (Sutskever et al., 2014) on a large scale of parallel corpora.

However, such approaches suffer from the *safe response* problem, that the decoder is easy to generate meaningless and generic responses such as "I think so" (Li et al., 2016). Previous studies have mainly tried in three typical ways to improve response specialty: (1) Modify objective functions to introduce diversity-promoting factors (Li et al., 2016; Li et al., 2017; Liu et al., 2018). (2) Encourage diversity of responses during beam search (Li et al., 2016; Vijayakumar et al., 2016), or introduce random factors by sampling (Adiwardana et al., 2020; Holtzman et al., 2019; Ippolito et al., 2019) rather than generate highest likelihood sentences. (3) Incorporate extra topic, cue-word, or style information to encourage informative responses (Xing et al., 2017; Yao et al., 2017; Mou et al.,2016; Gao et al., 2019; Wang et al., 2017). Despite various efforts to avoid safe responses, these approaches all generate responses in the monotonic left-to-right order.

We suspect that the monotonic left-to-right generation order, however, is not the most efficient way to discover informative responses, since it tends to generate high likelihood tokens at each step, typically lacking the guidance of a future specific goal. Though content-introducing approaches can introduce cue-words computed using either Pointwise Mutual Information (PMI) (Mou et al., 2016; Yao et al., 2017) or neural modules (Gao et al., 2019), these approaches rely on only one cue-word during decoding and do not explicitly model the linguistic intuition that the next token of the response is not only conditioned on the source sentence and previous response tokens but also the next point to express.

Hence, in this paper, we propose a novel but natural open-domain response generation approach, which first produces the most informative token as the cue-word, and then produces the next token under the guidance of the cue-word at each step, until the cue-word is produced as the token. We repeat the above process until reaching the end token. Figure 1 illustrates this generation approach. We name our approach as the Cue-Word Driven seq2seq model with a Shrinking Vocabulary (CWDSV).

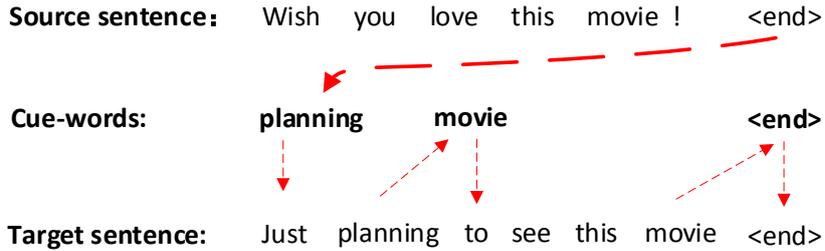

Figure 1: Cue-words driven generation process

Specifically, we use inverse document frequency (idf) (Jones, 1972) to measure the specialty of a token, and each cue-word determines a vocabulary made up of tokens whose idf value is less than the cue-word. This vocabulary is used by both the next cue-word and subsequent tokens driven by the current cue-word. Thus, the vocabulary shrinks during decoding, and when the end token is produced as the cue word, the vocabulary shrinks to size 1, since the idf value of the end token is 0.

We evaluate our model on the Sina Weibo short-text conversation dataset (Shang et al., 2015), and compare it with several strong baseline models. The experimental results show that the proposed CWDSV significantly outperforms the baseline models with much lower decoding complexity. Especially, our approach can converge to concrete semantics more efficiently during decoding.

## 2 Related Work

Our work is mostly related to three lines of research: (1) cue-word introducing neural response generation, (2) non-monotonic language generation. (3) language generation with dynamic vocabulary.

### 2.1 Cue-word Introducing Neural Response Generation

Cue word introducing approaches encourage informative responses by introducing a single cue word. Previous studies compute the cue word in two different ways. Mou et al. (2016) and Yao et al. (2017) calculate Pointwise Mutual Information (PMI) (Church et al., 1990) between source and target sentences, and then pick the cue-word with the highest PMI value when decoding. Such statistical methods, however, take the bag-of-words assumption, which ignores the word order. In contrast, Gao et al. (2019) introduce a discrete latent variable as the cue-word and compute it using neural components.

The above researches all introduce only one cue-word before decoding, and then use it to drive the entire beam search, which may restrict the decoder from exploring diverse responses. Unlike these researches, our approach produces the cue-word at the first step of decoding and then produces subsequent cue-words dynamically during decoding. In this way, the beam search can involve both cue-word and token generation, enabling the decoder to explore the full space of responses and discover subsequent informative cue-words throughout decoding.

### 2.2 Non-monotonic Language Generation

Though monotonic language generation has achieved enormous success (Bahdanau et al., 2014; Radford et al., 2019; Adiwardana et al., 2020), other work investigates different generation orders mainly in two styles: (1) the unconstrained generation order(Welleck et al., 2019; Emelianenko et al., 2019; Chan et al., 2019; Stern et al., 2019) and (2) the predefined generation order (Ford et al., 2018). We will only look back on the latter style, for it is more related to our approach.

Ford et al. (2018), inspired by the linguistic intuition that we may first think of points to express, and then serialize the whole sentence, try several different orders to generate sentences in a two-pass way, which fist generates a predefined part of words in the first pass, leaving other positions blank, and then fill in these blank positions in the second pass. They find that function-words-first and common-words first-orders outperform content-words-first and rare-words-first.

Though both use the predefined generation order, there are two main differences between our work and Ford et al. (2018): (1) We do not divide the vocabulary into two parts, e.g., common and rare words, or function and content words, but use a unified vocabulary and shrink it as cue-words are produced. (2) Our approach produces cue-words and tokens in one pass. Moreover, when producing cue words, our approach does not need to predict their positions, allowing the decoder to focus on cue-word semantics.

### 2.3 Language Generation with Dynamic Vocabularies

Due to Zif's Law (Zipf, 1949), the frequency of a word is inversely proportional to its frequency rank in the vocabulary. Hence, it is inefficient to use the entire vocabulary throughout decoding, especially when the vocabulary size is huge. To improve decoding efficiency, some studies in neural translation exploit a dynamic vocabulary that first select a subset of the entire vocabulary, and then use it to decode target sentences (Jean et al., 2014; L'Hostis et al., 2016; Shi and Knight, 2017; Mi et al., 2016).

Unlike the above researches, Wu et al. (2018) exploit a dynamic vocabulary in neural response generation, and jointly learn vocabulary construction and response generation. Their approach significantly outperforms baselines with less decoding time.

There are two fundamental differences in building the dynamic vocabulary between Wu et al.'s (2018) approach and ours: (1) They assume that the dynamic vocabulary obeys multivariate Bernoulli's distribution (Dai et al., 2013), i.e., occurrences of different words are identical. While in our approach, the vocabulary is conditioned on the last produced cue-word specialty, and our approach does not need hyper-parameter tuning of the vocabulary size. (2) They use the same vocabulary during decoding, but our vocabulary shrinks as cue-words are produced, reflecting the linguistic intuition that as the response nears the end, the semantics uncertainty shrinks, as well as the vocabulary.

## 3 Approaches

In this section, we will first formalize the problem and then describe the model in detail.

### 3.1 Problem Formalization

Suppose we have a dataset $D = \{(X_i, Y_i)\}_{i=1}^{N}$, where $Y_i$ is the response of $X_i$. We can formulate the generation probability of $Y_i$ as $p(Y_i|X_i) = \prod_{t=1}^{|Y_i|} p(y_t|y_{t-1}, \ldots, y_1, X_i)$. For each generation step, we divide $p(y_t|y_{t-1}, \ldots, y_1, X_i)$ into two sub-steps as follows:

$$p(y_t|y_{t-1}, \ldots, y_1, X_i) = p(y_t|k_t, y_{t-1}, \ldots, y_1, k_1, X_i) p(k_t|y_{t-1}, k_{t-1}, \ldots, y_1, k_1, X_i) \quad (1)$$

where $k_t$ is a cue-word and can be defined as the most informative token among $(y_j)_{j=t+1}^{|Y_i|}$. Formally, we define the specialty of a token as $I(y_j)$. As such, we can compute the cue-word $k_t$ as $k_t = \arg\max_{t+1 \leq j \leq |Y_i|} I(y_j)$. For each step, when producing $k_t(y_t)$, a specific vocabulary $V_t \subseteq V(V_t' \subseteq V)$, where $V$ is the entire vocabulary, is accordingly determined by $V$ and $k_{t-1}(k_t)$ as follows:

$$\begin{cases} V_1 = V \\ V_t = \{w_m, I(w_m) \leq I(k_{t-1}) \text{ and } w_m \in V\}, & (t > 1) \\ V_t' = V_{t+1} \end{cases} \quad (2)$$

Note that for cue-word generation, due to Formula 1, only a part of them need neural computation, when $k_t$ is to be re-predicted or $t = 1$, and otherwise $k_t$ can be set to $k_{t-1}$ directly.

Finally, our goal is to jointly learn cue-word and token generation tasks by maximizing log-likelihood as $\mathcal{L} = \sum_{i=1}^{N} \sum_{t=1}^{|Y_i|} \log(p(k_t|y_{t-1}, k_{t-1}, \ldots, y_1, k_1, X_i) + p(y_t|k_t, y_{t-1}, \ldots, y_1, k_1, X_i))$. In particular, the left half in the log can be omitted when $k_t = k_{t-1}$.

### 3.2 Cue-word Driven Seq2seq Model

Figure 2 illustrates the structure of our Cue-Word Driven seq2seq model with a Shrinking Vocabulary (CWDSV). CWDSV is based on an encoder-decoder framework with the attention mechanism. In detail, at each decoding step, CWDSV first estimates the probability of the cue-word $\widehat{k_t}$ and then passes $k_t$ to help estimates the token $y_t$ and to the next hidden state of the decoder to drive subsequent decoding. In detail, $k_t$ is the token with the maximum idf value among subsequent tokens. Accordingly, the vocabulary of the decoder shrinks during decoding as $k_t$ is re-predicted. Our model is compatible with various underlying networks such as Long-Short Term Memory (LSTM) (Hochreiter and Schmidhuber, 1997), Gated Recurrent Unit (GRU) (Cho et al., 2014) and Transformer (Vaswani et al., 2017). We use left-to-right LSTM for both encoder and decoder.

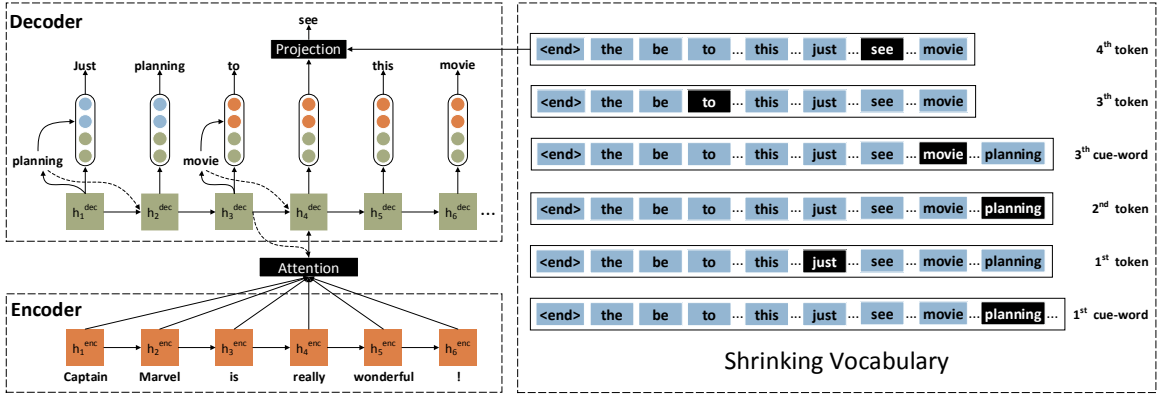

Figure 2: Architecture of the cue-word driven seq2seq model with a shrinking vocabulary.

Specifically, for the t-th decoding step, the decoder incorporates the last cue-word $k_{t-1}$, token $y_{t-1}$, and the context vector $c_t$ into the hidden state $h_t^{dec}$ by

$$h_t^{dec} = LSTM(h_{t-1}^{dec}, [k_{t-1}; y_{t-1}; c_t]) \quad (3)$$

where $[\cdot;\cdot;\cdot]$ means a concatenation of the three vectors. In detail, we use additive attention (Bahdanau et al., 2014) to produce $c_t$ as follows:

$$c_t = \sum_{s=1}^{|X|} \alpha_{t,s} h_s^{enc} \quad (4)$$

The attention score $\alpha_{t,s}$ is calculated as

$$\alpha_{t,s} = \frac{exp(e_{t,s})}{\sum_{s=1}^{|X|} exp(e_{t,s})} \quad (5)$$

$$e_{t,s} = v^T tanh(W_{att}[h_s^{enc}; h_{t-1}^{dec}]) \quad (6)$$

where $W_{att}$ and $v$ are parameters, and $[\cdot;\cdot]$ means a concatenation of the two vectors.

Then for decoding step t, CWDSV first estimates the generation probability of the cue-word by

$$\widehat{k_t} = l_k(y_{t-1}, c_t, h_t^{dec}, V_t) \quad (7)$$

where $\widehat{k_t}$ is a probability vector of size $|V_t|$. Then conditioned on $k_t$, we estimate the probability of the token as

$$\widehat{y_i} = l_y(k_t, y_{t-1}, c_t, h_t^{dec}, V_t') \quad (8)$$

where $\widehat{y_i}$ is a probability vector of size $|V_t'|$.

## 3.3 Shrinking Vocabulary

Besides the cue-word driven response generation process, another critical difference between CWDSV and the standard seq2seq model is that the decoding vocabulary naturally shrinks as cue-words are produced during decoding. We can efficiently implement this by ranking all tokens in the order of their idf values to make up the decoding vocabulary, and then use only a part of consecutive columns in the vocabulary matrix as the vocabulary shrinks.

## 3.4 Two-stage Beam Search

When decoding, we adapt beam search to our approach. As illustrated in Section 3.1 and 3.2, CWDSV produces the next token in two stages, i.e., (1) cue-word generation and (2) token generation. For each stage, k cue-words (or tokens) with the highest scores are picked and delivered to the next stage (or to the first stage of the next step), where k is the size of beam search. In particular, in the first stage, if the cue-word is not to be re-predicted, it is directly set to the previous one, and the probability of its belonging immediate hypothesis remains unchanged.

## 4 Experimental Settings

We evaluate the proposed CWDSV model on the Sina Weibo conversation dataset (Shang et al., 2015), and compare it with baselines. In this section, we will describe in detail the settings of our experiments.

### 4.1 Dataset

We conduct our experiments on the single-turn one-to-many short-text conversation dataset (Shang et al., 2015), which contains about 4.3 million post-response pairs from Sina Weibo, a twitter-like Chinese microblogging website. In detail, its training set contains 4244093 post-response pairs with 216889 posts, the validation set contains 19360 pairs with 997 posts, and the test set contains 3200 pairs with 800 posts, i.e., each post has four corresponding responses.

We employ pkuseg, a Chinese word segmenter (Luo et al., 2019), to tokenize posts and responses into Chinese word sequences. To deal with rare words, following Sennrich et al. (2016), we apply Byte Pair Encoding (BPE) (Gage, 1994) to split rare words into sub-words, and then use <UNK> to replace words with less than 40 occurrences, resulting in a vocabulary of size 49850, which covers 99.93% of words.

### 4.2 Implementation Details

We use a single-layer left-to-right LSTM for both the encoder and decoder as the base version of CWDSV. Additionally, considering the shrinking vocabulary size during decoding, to fairly compare with baseline models under a comparative decoding FLOPs, we also implement an enhanced version of CWDSV with a three-layer left-to-right LSTM decoder.

We set the dimension of all hidden vectors to 1024, and employ pretrained Chinese word embeddings released by Qiu et al. (2018) with dimension 300 and fine-tune them during the training stage. In addition, following Press and Wolf (2017), word embeddings are tied between the encoder and the decoder side. We train the model in mini-batches with the batch size 128. For parameter optimization, we use AdamW (Loshchilov and Hutter, 2017) optimizer with the learning rate of 1e-4 and the weight decay rate of 1e-7. All the above hyper-parameters are based on experience and are set as same for all comparison models. We stop training when the perplexity stops decreasing on the validation set.

At decoding time, for all models, we use beam search with the beam size of 5 and 10, respectively, to produce tokens with the highest likelihood, and then pick the hypothesis with the highest likelihood with length normalization. To alleviate the repetition problem in language generation, we use trigram beam blocking, which assumes that trigram rarely appears more than once in a sentence (Paulus et al., 2017). Our code is open-sourced and available at https://github.com/xxx/xxx.

### 4.3 Automatic Evaluation Metrics

Liu et al. (2016) have found that traditional automatic evaluation methods such as BLEU (Papineni et al., 2002) and embedding-based metrics have a weak correlation with human evaluation in conversation. One weakness of BLEU is that it requires exact word matching, regardless of synonymous words. Embedding-based metrics can capture synonymous words using vector similarity, but they ignore the word order, which is essential for sentence fluency.

BERT-score (Zhang et al., 2019) uses BERT (Devlin et al., 2018) to produce contextual word embeddings to compute similarity scores in semantics. Compared with traditional word embedding based automatic metrics, BERT-score can capture contextual semantics. Hence, we use BERT-score as the primary automatic metric to evaluate candidate sentences for all models. In detail, we pass sentences to BERT-base-Chinese (Devlin et al., 2018), enable the idf setting, which shows a better correlation with human evaluation (Zhang et al., 2019) and then use the last four layers to compute similarity scores. To compare sentence-level embeddings, following previous works (Wu et al., 2018; Liu et al., 2016), we also report *Embedding Average* and *Vector Extrema* as complementary metrics.

### 4.4 Human Evaluation Metric

We conduct 3-scale human annotation, recruiting three undergraduates who are all native speakers to evaluate the quality of generated responses. Responses from different models are randomly shuffled for each annotator. We use *Sensibleness and Specialty Average* (SSA) (Adiwardana et al., 2020) as the human annotation criteria. In detail, sensibleness evaluates whether a response is reasonable to the post, and specialty is to penalize safe responses. For example, "I don't know" is sensible to many posts, but would probably be penalized by specialty. Thus, the average of the two criteria rewards responses satisfying both sensibleness and specialty.

### 4.5 Comparison Models

We compare our model with the following three baseline models:

**S2SA**: This is the standard seq2seq model with the attention mechanism (Sutskever et al., 2014).

**S2SA-MMI:** This model uses MMI to re-rank results of beam search, and finally picks the response with the maximum $p(X|Y)$ (Li et al., 2016).

**S2SA-PMI**: This model uses the same structure with S2SA, and at each generation step, besides token and context embeddings, the cue-word embedding is fed into the decoder. Following Mou et al. (2016) and Yao et al. (2017), we compute the cue-word using PMI.

The above three baselines all use the single-layer left-to-right LSTM for both encoder and decoder and set hyper-parameters all the same as we have discussed in Section 4.2. Then two versions of our proposed models are listed as follows:

**CWDSV-base**: This is the proposed model which uses the single-layer left-to-right LSTM for both encoder and decoder.

**CWDSV-enhanced**: The shrinking vocabulary size during decoding may limit the representational capability of the decoder because the average amount of involved parameters and FLOPs of each step is significantly fewer than S2SA. To compensate for this gap, we simply use 3-layer stacked LSTMs for the decoder.

## 5 Experimental Results and Analysis

| Model | BERT-Score | | | | SentEmb | |
|---|---|---|---|---|---|---|
| | L=9 | L=10 | L=11 | L=12 | Avg | Ext |
| Beam=5 | | | | | | |
| S2SA | 57.86 | 63.26 | 66.80 | 61.20 | 72.42 | 71.27 |
| S2SA-MMI | 58.17 | 63.59 | 67.31 | 61.16 | 72.02 | 70.67 |
| S2SA-PMI | 57.23 | 62.88 | 66.58 | 59.62 | 71.21 | 69.97 |
| CWDSV-base | **58.55** | **63.88** | **67.46** | 61.39 | **74.94** | **73.15** |
| CWDSV-enhanced | 58.42 | 63.79 | **67.46** | **61.51** | 74.47 | 73.05 |
| Beam=10 | | | | | | |
| S2SA | 58.94 | 64.13 | 67.56 | 62.18 | 72.33 | 71.76 |
| S2SA-MMI | 59.41 | 64.64 | **68.28** | 62.24 | 71.76 | 70.39 |
| S2SA-PMI | 58.15 | 63.63 | 67.30 | 60.58 | 71.14 | 69.85 |
| CWDSV-base | **59.51** | **64.69** | 68.19 | **62.60** | **74.74** | **72.98** |
| CWDSV-enhanced | 59.27 | 64.48 | 68.09 | 62.53 | 74.46 | 72.74 |

Table 1: Automatic evaluation results of all compared approaches.

| Model | Human Evaluation | | | Kappa | |
|---|---|---|---|---|---|
| | Sens | Spec | SSA | Sens | Spec |
| S2SA | 0.58 | 0.54 | 0.56 | 0.68 | 0.58 |
| S2SA-MMI | 0.58 | 0.58 | 0.58 | 0.66 | 0.57 |
| S2SA-PMI | 0.48 | 0.47 | 0.47 | 0.61 | 0.56 |
| CWDSV-base | 0.69 | 0.65 | 0.67 | 0.61 | 0.56 |
| CWDSV-enhanced | **0.72** | **0.68** | **0.70** | 0.60 | 0.54 |

Table 2: Human evaluation results of all compared approaches.

In the following, we will report and analyze the experimental results and decoding complexity, conduct a case study on these results, and finally compare prediction accuracy between S2SA and CWDSV-base.

### 5.1 Automatic Evaluation Results

Table 1 shows the results with automatic evaluation results when the beam size is set to 5 and 10, respectively. The left-hand side shows BERT-score results, and under 7 out of 8 settings, our model achieves the highest BERT scores, except when Beam=5 and L=11, S2SA-MMI is rated the highest. Specifically, our models constantly outperform S2SA by 0.19 (Beam=5 and L=12) ~ 0.69 (Beam=5 and L=9) in all settings. The right-hand side shows sentence embedding similarities, namely, embedding average and vector extrema. Our approach outperforms all baselines in both two metrics.

The results in Table 1 suggest that the responses generated by our model have the highest semantic similarity with human-produced sentences. We also note that in most settings, the score of CWDSV-base is slightly higher than CWDSV-enhance, but combined with human evaluation we will discuss in Section 5.2, CWDSV-enhanced indeed generates better responses.

Turning to baseline models, S2SA-MMI outperforms S2SA in 7 out of 8 settings of the BERT-score. This may suggest that BERT-Score can penalize safe responses, for safe responses probably have low semantic similarity with selected human-produced sentences. In addition, S2SA-PMI achieves lowest BERT scores and sentence embedding similarities among all models, suggesting that cue-words calculated using PMI do not necessarily lead to higher semantics similarity with human-produced sentences.

### 5.2 Human Evaluation Results

Table 2 gives the results using human evaluation. We conduct a 3-scale human evaluation only under the beam size of 10 because of the economic cost. CWDSV-base achieves significantly higher scores on both sensibleness and specialty than all the baselines., and CWDSV-enhanced achieves the highest scores. Note that though using additional decoder layers, the decoding complexity of CWDS-enhanced is much lower than S2SA, as we will discuss in Section 5.4. S2SA-MMI achieves comparative sensibleness with S2SA, but higher specialty, at the expense of significantly higher decoding complexity for computing $p(X|Y)$ for each hypothesis.

We also report Fleiss' Kappa (Fleiss and Cohen, 1973) for all models on the right-hand side of Table 2 on sensibleness and specialty, respectively. The kappa value of sensibleness ranges at 0.60~0.68, which indicates substantial agreement of different annotators. Specialty's kappa value is a little lower, ranging at 0.54~0.58, indicating moderate agreement. Due to our observation of the controversial cases in terms of specialty, they often express some emotion but may not be very specific, such as "that's funny".

### 5.3 Case Study

Figure 3 shows four example posts with responses generated by different models under discussion. S2SA tends to generate safe responses or even repeat the post. S2SA-MMI can generate responses of a higher specialty. In some cases, however, it can generate similar responses with S2SA. Moreover, it can lead to low fluency responses if such responses appear in beam search hypotheses and are accidentally evaluated with the highest $p(X|Y)$. S2SA-PMI can produce more specific information using the cue-word, but the cue-word calculated using PMI is not always appropriate. CWDSV-base constantly tries to produce high related responses due to its cue-word driven generation process, but in some cases, it can produce low fluency responses, partly because of its limited capacity in the decoder side. CWDSV-enhanced can generate high related responses with better fluency, due to its extra decoder layers.

| Post | 这些英文千万不要不懂装懂。<br>Don't pretend to know these English. | 麦当劳良心早餐大幅涨价, 哎<br>McDonald's has dramatically increased the price of breakfast, oh no |
|---|---|---|
| S2SA | 千万不要不懂装懂, 千万别不懂装懂。<br>Don't pretend to know what you don't. Don't pretend to know what you don't. | 麦当劳也涨价了吗?<br>Has McDonald's raised its price too? |
| S2SA-MMI | 千万不要不懂装懂, 千万别不懂装懂,<br>Don't pretend to know what you don't. Don't pretend to know what you don't. | 麦当劳也涨价了吧。<br>McDonald's has raised its prices, too. |
| S2SA-PMI | 英文不好, 看不懂。<br>I can't understand English well. | 麦当劳和麦当劳有什么区别?<br>What's the difference between McDonald's and McDonald's? |
| CWDSV-base | 这个要好好学习一下。<br>I've got to learn this. | 我也想去吃麦当劳了。<br>I want to eat McDonald's, too. |
| CWDSV-enhanced | 不懂英文的人伤不起啊!<br>It's so sad that I don't understand English! | 麦当劳的良心大大地坏了<br>McDonald's conscience is rotten |
| Post | 泡沫沾彩色甲油, 漂亮吧。<br>Foam board with colored nail polish. Nice. | 公司某员工得 2013 年工作计划。<br>This is the 2013 work plan of an employee of the company. |
| S2SA | 好漂亮的指甲油啊!<br>What a beautiful nail polish! | 公司的员工都是这么想的。<br>That's what the company's employees all think |
| S2SA-MMI | 好漂亮的指甲油, 好漂亮<br>What a beautiful nail polish. So beautiful | 公司的员工都是工作人员<br>The company's employees are all staff |
| S2SA-PMI | 好漂亮, 好漂亮的泡泡<br>So Beautiful! What beautiful bubbles | 这样的员工不容易啊!<br>Such employees are not easy! |
| CWDSV-base | 我想知道是怎么涂的?<br>I wonder how it's painted? | 什么时候我们也能有这样的公司呢?<br>When will we have a company like this? |
| CWDSV-enhanced | 好漂亮啊, 怎么涂的?<br>That's beautiful. How do you paint it? | 有这样的员工真是太幸福了。<br>It's so happy to have such employees. |

Figure 3: Examples of generated responses by different models

### 5.4 Decoding Complexity

We compare the decoding complexity of different models using the following two metrics:

**Floating Point Operations (FLOPs)**: At decoding time, we record FLOPs for each model as one of decoding complexity metrics.

**Activations:** Recent work has found that compared with FLOPs, activations, namely the size of tensors in the neural network, has a more significant influence on run time on memory-bound hardware accelerators such as GPU and TPU (Radosavovic et al., 2020). According to language generation network structures, we define network activations as the size of output tensors of all arithmetic operations (e.g., matrix multiplication, tanh, and sigmoid).

We show the metric results of different beam sizes in Figure 4. CWDSV-base needs fewer FLOPs and significantly fewer activations than S2SA. The decreased complexity is due to the shrinking vocabulary size during decoding. Moreover, CWDSV-enhanced, though use 3-layer left-to-right LSTM in the decoder, needs comparative FLOPs and significantly fewer activations than S2SA.

### 5.5 Prediction Accuracy

We compare the prediction accuracy of different steps of S2SA and CWDSV-base to see how cue-word prediction influence the decoding process. Specifically, for our models, when a decoding step contains both cue-word and token prediction, it is treated as correct if and only if both the cue-word and the token are predicted correctly. Furthermore, we also compare the rate of whether the top-5 candidate tokens contain the reference token since there is generally no standard answer in conversation.

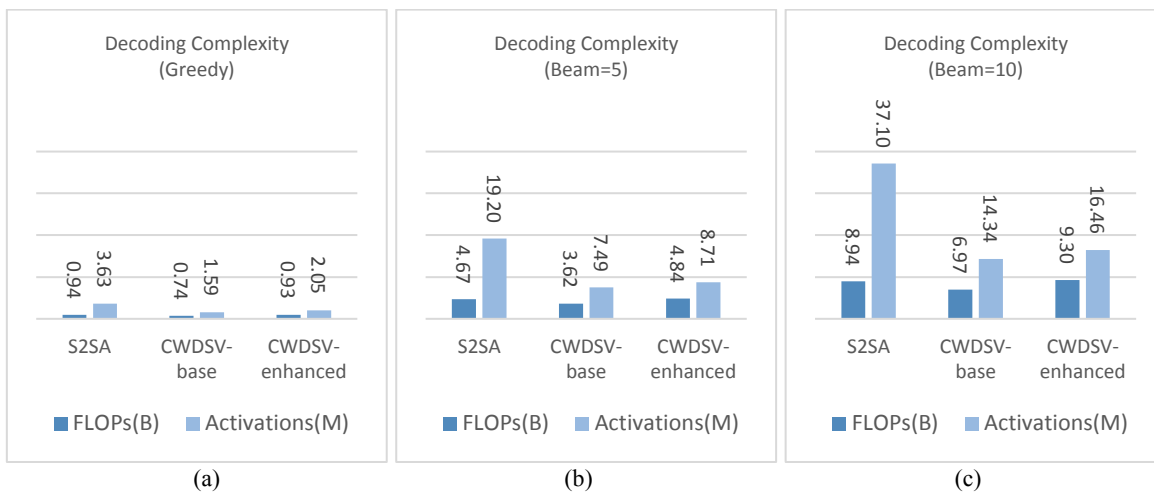

Figure 4: Decoding complexity

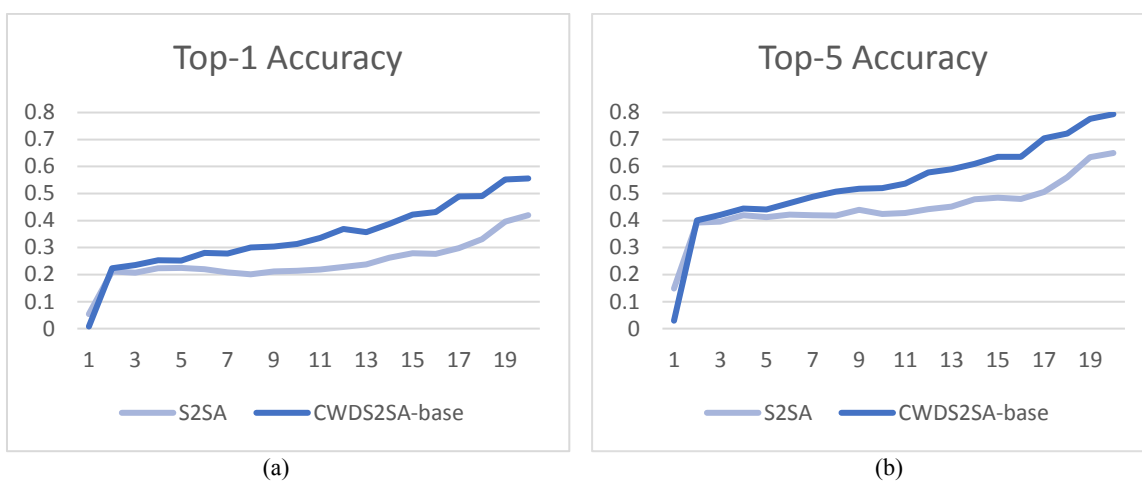

Figure 5: Prediction accuracy and contained rate of different steps

Figure 5 shows the top-1 and top-5 prediction accuracy on the test set of 3200 reference sentences. CWDSV-base achieves higher accuracy at most steps, except 0.78% (top-1) and 2.94% (top-5) at the first step, compared with 5.38% (top-1) and 14.84% (top-5) of S2SA, for CWDSV-base needs to predict both the first cue-word and token. Moreover, compared with S2SA, CWDSV-base shows a stable upward trend from step 2 to 8. This suggests that it can have more information gain per step, especially in the first few steps, and converge to specific semantics more efficiently, which should help generate informative responses.

## 6 Conclusion and Future Work

To alleviate the notorious safe response problem in open domain response generation, and to generate informative responses efficiently, we propose a cue-word driven model with a shrinking vocabulary during decoding. We evaluate our model on the Sina Weibo short-text conversation dataset (Shang et al., 2015). The experimental results show that our approach significantly outperforms strong baseline models, with much lower decoding complexity.

We only apply our approach to response generation in the present study. However, it has the potential to be applied in various open-ended text generation tasks such as story generation (Kybartas and Bidarra, 2016) and contextual text continuation (Radford et al., 2019), in that these tasks have a similar linguistic intuition. We leave it for future study.